\documentclass[times,twocolumn,final]{elsarticle}

\usepackage{ipcai_arxiv}
\usepackage{graphicx}%
\usepackage{multirow}%
\usepackage{amsmath,amssymb,amsfonts}%
\usepackage{amsthm}%
\usepackage{mathrsfs}%
\usepackage[title]{appendix}%
\usepackage{xcolor}%
\usepackage{textcomp}%
\usepackage{manyfoot}%
\usepackage{booktabs}%
\usepackage{algorithm}%
\usepackage{algorithmicx}%
\usepackage{algpseudocode}%
\usepackage{listings}%
\usepackage{multirow}
\usepackage{pifont}%
\usepackage{adjustbox}
\usepackage{graphicx}
\usepackage{hyperref}
\usepackage{float}
\usepackage{placeins}
\usepackage{fancyhdr}

\begin{document}
\fancypagestyle{firstpagestyle}{
    \fancyhf{} 
    \renewcommand{\headrulewidth}{0pt} 
    \fancyhead{} 
    \fancyfoot{} 
    \fancyhead[CO]{\em \fontsize{9pt}{8pt}\selectfont This article has been accepted to the International Conference on Information Processing in Computer-Assisted Interventions, 2024.}
}

\fancypagestyle{default}{
    \fancyhf{}
    \fancyhead[R]{\thepage} 
    \renewcommand{\headrulewidth}{0.4pt} 
    \fancyhead[LO]{\em \fontsize{9pt}{8pt}\selectfont This article has been accepted to the International Conference on Information Processing in Computer-Assisted Interventions, 2024.}
}



\title{On-the-Fly Point Annotation for Fast Medical Video Labeling}

\author[1]{Adrien \snm{Meyer} \fnref{corresp}}
\author[1,2]{Jean-Paul \snm{Mazellier}}
\fntext[corresp]{Corresponding author: \texttt{meyer.adrien1996@gmail.com}}
\author[2,3]{Jérémy \snm{Dana}}
\author[1,2]{Nicolas \snm{Padoy}}

\address[1]{ICube, University of Strasbourg, CNRS, France}
\address[2]{IHU Strasbourg, France}
\address[3]{Department of Diagnostic Radiology, McGill University, Montréal, Canada}

\received{XXX}
\finalform{XXX}
\accepted{XXX}
\availableonline{XXX}
\communicated{XXX}

\begin{abstract}
\textbf{Purpose}:
In medical research, deep learning models rely on high-quality annotated data, a process often laborious and time-consuming. This is particularly true for detection tasks where bounding box annotations are required. The need to adjust two corners makes the process inherently frame-by-frame. Given the scarcity of experts' time, efficient annotation methods suitable for clinicians are needed.

\noindent\textbf{Methods}:
We propose an \textit{on-the-fly} method for \textit{live video annotation} to enhance the annotation efficiency. In this approach, a continuous single-point annotation is maintained by keeping the cursor on the object in a live video, mitigating the need for tedious pausing and repetitive navigation inherent in traditional annotation methods. This novel annotation paradigm inherits the point annotation’s ability to generate pseudo-labels using a point-to-box teacher model. We empirically evaluate this approach by developing a dataset and comparing on-the-fly annotation time against traditional annotation method.

\noindent\textbf{Results}:
Using our method, annotation speed was $3.2\times$ faster than the traditional annotation technique. We achieved a mean improvement of \(6.51 \pm 0.98\) AP@50 over conventional method at equivalent annotation budgets on the developed dataset.

\noindent\textbf{Conclusion}:
Without bells and whistles, our approach offers a significant speed-up in annotation tasks. It can be easily implemented on any annotation platform to accelerate the integration of deep learning in video-based medical research.
\\

\end{abstract}

\maketitle
\thispagestyle{firstpagestyle}

\section{Introduction}\label{intro}

Object detection is increasingly recognized as a critical tool in medical video analysis, with applications ranging from identifying landmarks, organs and lesions to tracking surgical instruments in real-time procedures to assess safety \cite{AIinMed,mascagni2022computer}.
Yet, while modern deep learning-based object detectors have shown notable success, their effectiveness is largely dependent on the availability of extensive annotated data \cite{dino, rtmdet, polipReview}. This becomes particularly challenging in the medical domain, given the labor-intensive nature of annotation and the constraints on experts' time, especially given their primary clinical duties. Currently, the process of video object annotation is frame-based, which is suboptimal and highly time-consuming, even with the use of interpolation tools. As a result, most studies rarely annotate more than one or two hundred videos \cite{twinanda2016endonet,srivastav2018mvor}. This methodological constraint hampers the democratization and scalability of deep learning in medical procedures, as there is a compelling requirement for the collection and annotation of diverse, multicentre, and multi-operator data.

To address these challenges in domains where the opportunity cost of expert time is high, methods to accelerate the annotation process generally fall into one of two non-exclusive categories:
\begin{enumerate}
\item Methods to scale the pool of available annotators by lowering the expertise barrier. For instance, experts could annotate keyframes that non-experts, possibly crowdsourced contributors, use as references to expand the dataset \cite{fastMLannotationMed}. However, it requires sufficient resources for managing human capital at scale. The gains in annotation production must outweigh the overheads of annotators training, follow-up of annotation quality and project management.
\item Methods to lessen the annotation burden per annotator through the use of weaker localization labels, such as absence/presence, gaze, scribbles, or points \cite{SAM,pointDetr}. The weak labels are used in a (Weakly) Semi-Supervised Object Detection ((W)SSOD) pipeline, which uses a small set of box-level labeled images as well as a larger set of weakly labeled images to train detectors. In Point-DETR \cite{pointDetr} and Group R-CNN \cite{grouprcnn}, the authors propose to learn a point-to-box teacher model to generate pseudo-box labels.
\end{enumerate}

We position our work on the later categories. We propose a new annotation paradigm focused on 
\textit{live video annotation}. Our proposal is a shift from the conventional \textit{frame-based} approach to a more dynamic \textit{video-based point annotation} strategy. Unlike bounding boxes, point annotations have only two degrees of freedom ($x$ and $y$ coordinates) and can be adjusted with a single drag of a pointer (i.e., mousepad, pencil on tactile screen or eye gaze). Leveraging this inherent property, we introduce \textit{on-the-fly} annotation, where a 2D cursor enables continuous tracking on \textit{live video}, producing a point annotation similar to a temporal scribble (see Fig.~\ref{figOverview}b). This eliminates frame-by-frame annotation, speeding up the process while maintaining the advantages of point-based WSSOD. We leverage point-to-box teacher models \cite{pointDetr,grouprcnn} for the generation of pseudo-box labels derived from the point annotations to train detectors.


To validate our annotation method, we constructed a liver ultrasound video dataset and compare the efficiency of standard bounding box and on-the-fly point annotation (OTF) methods. Given the high level of expertise required in the ultrasound domain, enhancing the annotation efficiency of experts is crucial. With varying annotation budgets, we train two type of teacher models: one using point-to-box models in a WSSOD paradigm with OTF and the other as a traditional object detector. These teachers generate pseudo-labels for training student models. Unlike methods utilizing interpolation or crowdsourcing, our approach ensures that every  frame in the video is weakly-annotated by the experts.

Our contributions are twofold: (1) We introduce the novel task of live video annotation, and (2) we present an on-the-fly point annotation method optimized for this task within a WSSOD framework.

\section{Related work}\label{relatedWork}

\subsection{Crowdsourcing Annotations}
Crowdsourcing aims to match expert annotation performance in tasks like image annotation. In healthcare domains, the study by \cite{rother2021assessing} on hepatic steatosis reveals that crowdworker annotation reliability isn't guaranteed by annotator certainty or agreement, yet a larger crowd slightly outperformed a few experts. The study by \cite{heim2018large} suggests that crowds can refine automatic 3D segmentation of liver CT scans to a level comparable to experts, although at a slower pace. Crowdsourcing can offer scalability in annotations but requires substantial setup that becomes cost-effective only at large project scales. Its suitability varies by project and modality and may underperform in tasks demanding high expertise.

\subsection{Semi-Supervised/Weakly Supervised Object Detection}
Semi-Supervised Object Detection (SSOD) and Weakly-Supervised Object Detection (WSOD) aim to mitigate the high cost of data annotation. 
SSOD methods leverage a mix of a few box-level labeled images and many unlabeled ones with two main approaches; consistency regularization techniques, to stabilize the detector's predictions across variably augmented images \cite{jeong2019consistency}, and pseudo-labeling, where a teacher model trains on labeled data to generate pseudo-labels for unlabeled data. A student model then trains on both datasets for improved performance \cite{liu2021unbiased,wang2021data}.
WSOD methods use abundant but weakly annotated data, such as image labels \cite{wsdetr,pcl}. The studies by \cite{vardazaryan2018weakly} and \cite{kim2021weakly} utilize class activation maps in WSOD methods to enable both detection and localization of surgical tools in endoscopic videos and breast cancer in ultrasound images respectively, without spatial annotations.

Combining these approaches, Weakly Semi-Supervised Object Detection (WSSOD) methods use both box-level and weakly labeled images to train detectors, aiming to propose a favorable trade-off between annotations cost and performances. \cite{ouyang2023weakly} detect lung consolidations in ultrasound videos, using video-level labels (presence in at least one frame) and a teacher-student training strategy.


\section{Methodology}\label{method}
In this section, we first introduce the task of live video annotation and discuss why bounding box annotation is suboptimal from a video annotation point of view. Next, in order to address it, we illustrate our novel on-the-fly point annotation as an efficient alternative.

\subsection{Live Video Annotation}

\begin{figure*}[h!]%
\centering
\includegraphics[width=\linewidth]{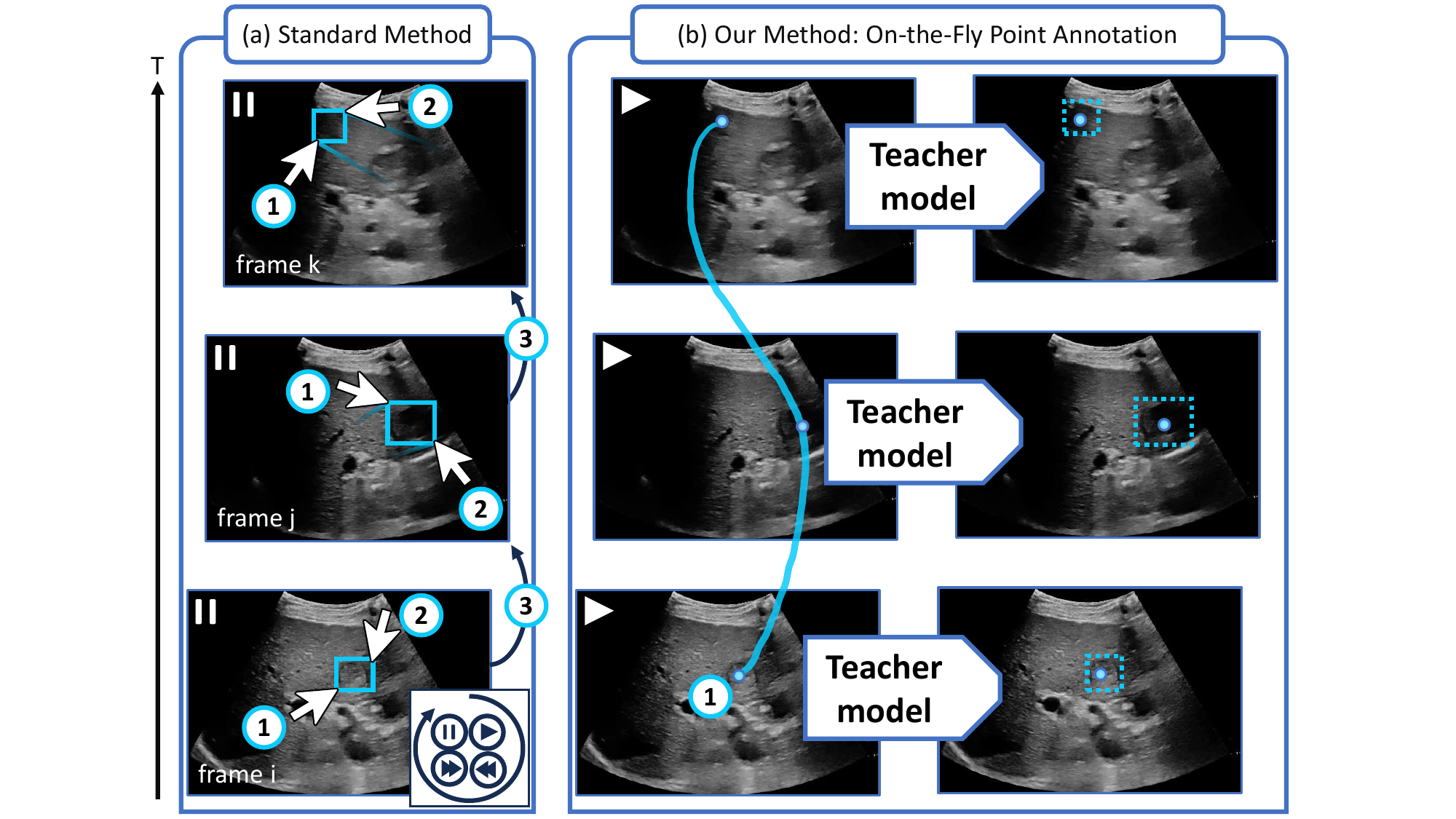}
\caption{(a) Conventional bounding box annotation approach on static frames. \ding{172} \ding{173} adjust the two corners, \ding{174} video navigation; (b) Our proposed \textit{on-the-fly} point annotation method on live video. \ding{172} pointing of the targeted structure. Box cyan - lesion; solid lines - ground truth; dashed lines - predictive pseudo labels.}\label{figOverview}
\end{figure*}

We introduce Live Video Annotation as a new approach to streamline the process of object annotation in videos. The key goal of live video annotation is to reduce or completely eliminate the need to frequently pause the video, thereby making the annotation process more fluid and efficient. This raises a critical question: How can a dataset be densely annotated with spatial instance information (Fig.~\ref{figOverview}a) without substantial pausing?

\thispagestyle{default}
Considering the standard annotation pipeline as performed on a dedicated video annotation software with keyframes interpolation (Fig.~\ref{figOverview}a), the annotation process is as follows: i) clicking on the two corners of a tight box around the object \ding{172} \ding{173}, ii) navigating the video with play/forward/backward \ding{174}, iii) pause on the next keyframe, iv) go back to step (i). These convoluted steps are the result of the multi-click nature of boundary annotation, which is not compatible with a continuous annotation on a streamed video. In this paper, the aforementioned method will be referred to as the 'BBox method' for ease of discussion and simplicity. From those observations, we propose to use a weaker localisation label such as the point, which only requires one click/drag to adjust its position over a continuous video playback.

\subsection{On-the-fly Point Annotation}

We propose a novel \textit{on-the-fly} point annotation (OTF) strategy for video streaming. In this scenario (Fig.~\ref{figOverview}b), the user is asked to continuously point at the targeted structure during the video playback \ding{172}, reducing the tedious pausing and back-and-forth video navigation associated with the standard annotation method (Fig.~\ref{figOverview}a). In practice, the annotator still needed to pause occasionally for video understanding, taking breaks, or stopping the live annotation when the object disappeared from view. In videos where objects frequently appear and disappear, continuously stopping OTF annotations can be inefficient as it interrupts the live annotation process to precisely find the frames where the object is not visible. A smoother, less conservative approach could involve performing annotations purely on-the-fly, without pausing the video, thereby enhancing workflow fluidity and reducing annotation time. However, to maintain the quality of annotations and mitigate the risk of introducing false positives, it might be prudent to exclude annotations at the temporal edges corresponding to the annotation stoppages. In our experiment, we adopt the more conservative approach of stopping the video and precisely stopping the annotation when the object disappears. In the annotation process for our study, we utilized a reduced playback speed of 0.2x. This slower speed was essential to accurately track rapid changes in the videos, which are difficult to observe at the normal speed (1x). This adjustment helped minimize potential errors in annotation. However, we recognize that this method may not be universally applicable, particularly in scenarios involving faster movements, and the choice of playback speed might need to be tailored to the specific dataset being annotated. The resulting annotation maintains the advantages of point-based WSSOD, i.e. Point-DETR \cite{pointDetr} and Group R-CNN \cite{grouprcnn}.

We adopt the self-training pipeline of \cite{pointDetr}. Given a small number of supervised images and a large number of weakly supervised images: i) Train a teacher model on available labeled images, ii) Generate pseudo-labels of weakly OTF annotated images using the trained teacher model and iii) Train a student model with fully labeled images and pseudo-labeled images.\\
\thispagestyle{default}

To verify the benefit of the proposed OTF, we used a  dataset named STARHE of liver ultrasound videos, as described below, with annotated lesions with both bounding box annotation and OTF. We timed both annotation methods on subsets of the annotated videos to compare the annotation speed. Next, we studied whether the points annotated using OTF accurately lie within the corresponding bounding boxes. This assessment allowed us to determine the consistency of the OTF points and whether they accurately tracked and moved in alignment with the objects being annotated. Finally, we conducted comparative studies between two self training scenario $S_{OTF}$ and $S_{BBox}$. The first leveraged OTF while the latter did not. This comparison aimed to assess the compatibility and effectiveness of OTF-based pseudo labels within a WSSOD pipeline. 
%

\begin{figure*}[h!]%
\centering
\includegraphics[width=\linewidth]{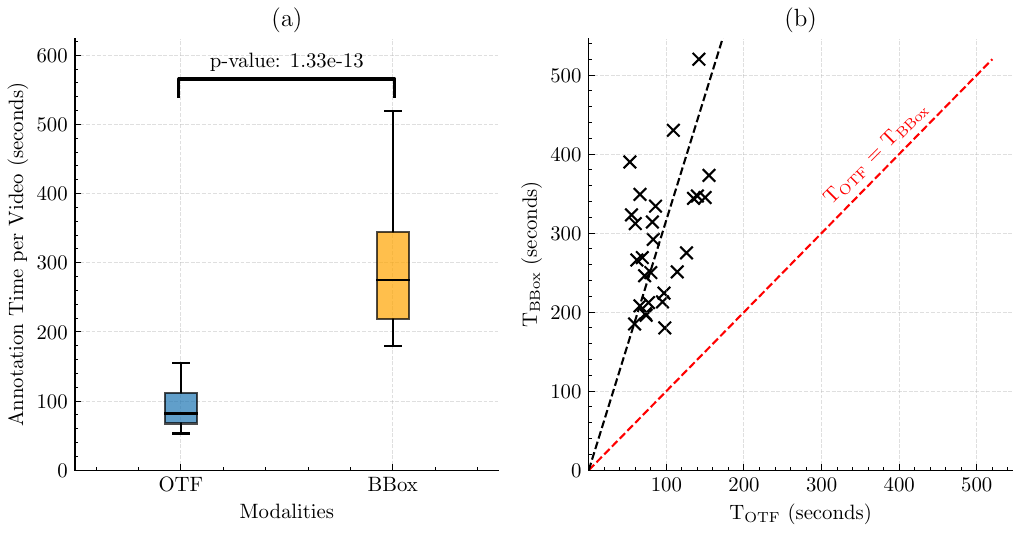}
\caption{(a) Box plot comparing annotation times between OTF and BBox method. (b) Pairwise comparison of annotation times for each timed video, with a fitted line illustrating the relationship between $T_{BBox}$ and $T_{OTF}$.}\label{timeGain}
\end{figure*}
We employ DETR \cite{carion2020end} and Faster R-CNN \cite{ren2015faster} as our student models. DETR uses the transformer architecture to simplify object detection, removing hand-crafted elements like non-maximum suppression and anchor generation, while maintaining performance on par with Faster RCNN \cite{ren2015faster}.
We employ Point-DETR and Group R-CNN as a teacher models in \(S_{OTF}\). Point-DETR extends DETR by incorporating both images and point annotations as inputs. It employs a point encoder to map these point annotations to object queries, enhancing detection performance through strong prior localization and class. Group R-CNN \cite{grouprcnn}, building on classic R-CNN architecture, introduces
instance-level proposal grouping and assignment, coupled with instance-aware representation learning, to effectively translate point annotations into precise box proposals. Those models are trained on a box-level annotated video set \(Box(S_{OTF})\), and subsequently used to create pseudo labels on the weakly labeled videos \(OTF(S_{OTF})\). Therefore, the corresponding annotation budget \(B_{OTF}\) can be expressed as
\begin{equation}
B_{OTF} = T_{Box} \times \left|Box(S_{OTF}) \right| + T_{OTF} \times \left| OTF(S_{OTF}) \right|
\label{eq1}
\end{equation}
where \(T_{BBox}\) and \(T_{OTF}\) represent the average annotation times required to annotate a video using the bounding box or OTF method, respectively. $\left|Box(S_{OTF}) \right|$ and $ \left| OTF(S_{OTF}) \right|$ are the number of annotated videos with BBox or OTF method for the scenario \(S_{OTF}\), respectively.
For \(S_{BBox}\), which utilizes the classic DETR as a teacher model, the annotation budget \(B_{BBox}\) is calculated as
\begin{equation}
B_{BBox} = T_{BBox} \times \left|Box(S_{BBox}) \right|
\label{eq2}
\end{equation}

Since no annotation is required during inference, this model budget solely depends on the BBox method. To ensure a fair comparison between the two models, we increase the number of annotated videos $\left|Box(S_{BBox}) \right|$ such that \(B_{BBox} = B_{OTF}\). In this way, the time spent on weak annotations is effectively converted into additional box-level annotated videos.

\section{Experimental Setup}\label{xpSetup}
\thispagestyle{default}
\subsection{STARHE Dataset}

We developed and tested our OTF method using a newly created dataset, STARHE (Risk Stratification of Hepatocarcinogenesis), registered at ClinicalTrials.gov (Identifier: NCT04802954). This dataset gathered liver ultrasound videos acquired using a standardized protocol. The current hepatocellular carcinoma (HCC) screening program in France relies on biannual liver ultrasound. However, the performance of this screening program is poor which can be explained by poor liver visualisation using ultrasound in some patients (e.g., obesity, steatosis, ...), operator dependency, or limited patient compliance. Given the anticipated surge in HCC-related mortalities by 2030, a more effective screening strategy is needed. Our aim is to develop an automated method for detecting HCC lesions during ultrasound screenings, thereby minimizing missed lesions and delays in diagnosis.

In our study, an experienced  clinician (radiologist) annotated a set of 125 ultrasound videos with dedicated video annotation software with interpolation tools, employing both OTF and BBox annotation methods for each video, as shown in Fig. \ref{figOverview}. The annotations were performed using a mouse cursor. A minimum one-month interval was maintained between each annotation type to ensure no recall bias, with previous annotations being hidden during the subsequent session. For a subset of 27 videos, the annotation process was timed to compare the efficiency of both methods. Specifically, in $S_{OTF}$, annotations were made on live videos played at 0.2x speed, ensuring a comprehensive understanding of the video content throughout the annotation process. The videos in our dataset had a duration of 10 seconds. We partitioned our dataset as follows: 20\% for testing (25 videos), 10\% for validation (13 videos), and 70\% for training (87 videos). Within the training set, we further divided the data into a box-level annotated set and a weakly annotated set to conduct experiments with varying annotation budgets. Initially, the division was set at 20\% box-level annotated and 80\% weakly annotated. We then incrementally transferred 5\% from the weakly annotated set to the box-level annotated set, until the distribution reached 60\% box-level annotated and 40\% weakly annotated. We report the average precision at an intersection over union threshold of 0.5 (AP@50), averaged over 3 runs with random data splitting.



\subsection{Teacher models Training}
In \(S_{OTF}\), we use Point-DETR and Group R-CNN as our teacher models. We pretrain the models on $20\%$ of the COCO dataset \cite{coco}, limited by computational budget constraints, preventing the use of the full dataset. 70$k$ iterations of AdamW training with mini-batch 4 on $2\times$ Nvidia V100 are performed for fine-tuning on our STARHE dataset, using an initial learning rate of 1e\textsuperscript{-4}. We divide the learning rate by 10 at $50k$ iterations. We use random flipping, resizing and random croping as data augmentation. In our experiments, we train the teacher models with noise on the point up to $25\%$ of its respective box dimension. Note that point noising serve as a data augmentation and is not used during inference. To reduce frame redundancy, we trained the models using every eighth frame from the annotated videos, approximately equating to 2.5 frames per second. Our implementation is based on the MMDetection library~\cite{mmdetection}. Similarly, in \(S_{BBox}\), we employ DETR as our baseline teacher model. In both \(S_{OTF}\) and \(S_{BBox}\), we use DETR and Faster R-CNN as our student models. The student models utilizes pretrained weights from the entirety of the COCO dataset.

\section{Results}\label{Results}

In our initial experiment, we examined the annotation speed between the OTF and BBox methods. As illustrated in Fig.\ref{timeGain} (a), we observed a statistically significant acceleration, with the OTF method being on average 3.2 times faster than the BBox method (\(p = 1.33e^{-13}\)). We present a pairwise comparison of annotation times for each video in Fig.\ref{timeGain} (b), illustrating a distinct relationship between \(T_{BBox}\) and \(T_{OTF}\). This visualization highlights a consistent pattern: the longer a video takes to annotate using one method, the longer it tends to take using the other method as well (see the fitted line). This suggests a consistent difficulty level in video annotations, such as lesion visualisation due to poor conspicuity, irrespective of the method used. Still, the OTF method consistently proves to be significantly faster than the BBox method in our annotation scenarios.

\begin{figure}%
\centering
\includegraphics[width=\linewidth]{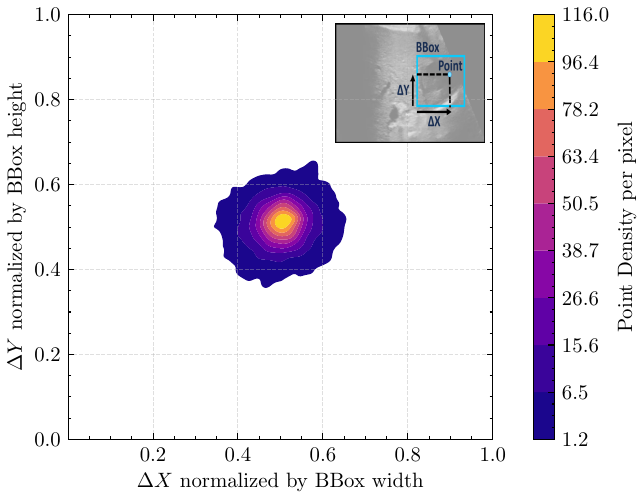}
\caption{Spatial density of OTF annotation locations across all videos with respect to corresponding boxes. The x-axis represents the horizontal position, and the y-axis represents the vertical position of the OTF within the box. Yellow areas indicate regions with a higher density of annotations, while dark blues indicate a lower density.}\label{pointsDistribution}
\end{figure}

To better understand the distribution and localization of annotated points with respect to corresponding BBox, we employ a kernel density estimation plot as illustrated in Fig.\ref{pointsDistribution}. This allows us to present the data as a heatmap. The axes, ranging from 0 to 1, represent the relative dimensions of the boxes, with values indicating the normalized position of corresponding OTF annotation within them. Areas with a higher color intensity signify regions with a denser concentration of annotations. Two key observations are made regarding the OTF annotations. First, all OTF annotations systematically fall within their respective boxes, confirming the precision of this annotation method. Secondly, a significant concentration of OTF annotations is observed around the near-center regions of the boxes, despite the annotator not being explicitly instructed to target the centre of the structures. This denotes that the OTF method effectively facilitates accurate tracking of the anatomical structures in question. This tendency towards centre-annotation could potentially be an instinctive approach adopted when tracking oval-like structures, such as lesions, enabling a more intuitive annotation process.

\thispagestyle{default}


\begin{figure*}[h!]%
\centering
\includegraphics[width=\linewidth]{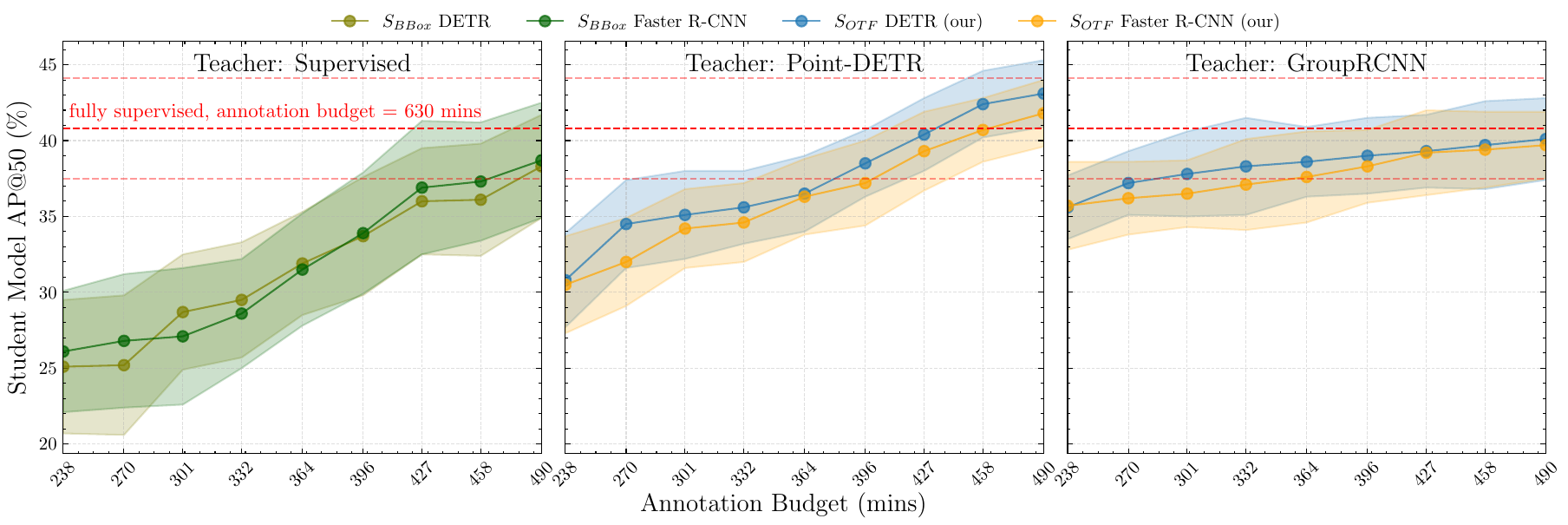}
\caption{AP@50 of Student models under similar annotation budgets, utilizing a blend of box-level and pseudo labels. Results, along with the standard deviation, are computed based on three individual runs. $S_{OTF}$ pseudo labels are from point-to-box models using OTF annotation, whereas $S_{BBox}$ uses Faster R-CNN or DETR-derived pseudo labels without prior.}\label{perf_students}
\end{figure*}

Finally we investigated the integration of OTF labels into a WSSOD pipeline. A comparative study was conducted between the two self-training scenarios, \(S_{OTF}\) and \(S_{BBox}\), to evaluate the effectiveness of our method in generating pseudo-labels for downstream applications. We report the AP@50 and standard deviation, calculated over three runs, with equivalent annotation budget in Fig.\ref{perf_students}. \(S_{OTF}\) achieves a mean improvement of \(6.51 \pm 0.98\) AP@50 over \(S_{BBox}\). Using Group R-CNN, we achieved better results than Point-DETR, especially in scenarios with smaller annotation budgets. With a 238-minute annotation budget, Group R-CNN combined with DETR achieved an AP@50 of 35.7\%, and 35.6\% when paired with Faster-RCNN. This success is due to Group R-CNN's multi-scale approach and CNNs' efficiency with limited data. Interestingly, \(S_{OTF}\) even surpasses the performance of the fully supervised scenario, which we infer to be a consequence of a label smoothing effect induced during the pseudo-label generation process which aligns more closely with the expectations of the student model, facilitating more effective learning and acting as a regularization mechanism. While $S_{OTF}$ exceeds the fully supervised model's performance for annotations budget over 427 minutes, it remains within the error margin of the supervised model.
 Overall, we achieve the same performance as the fully supervised baseline with $68\%$ of its annotation budget.

\begin{figure}%
\centering
\includegraphics[width=\linewidth]{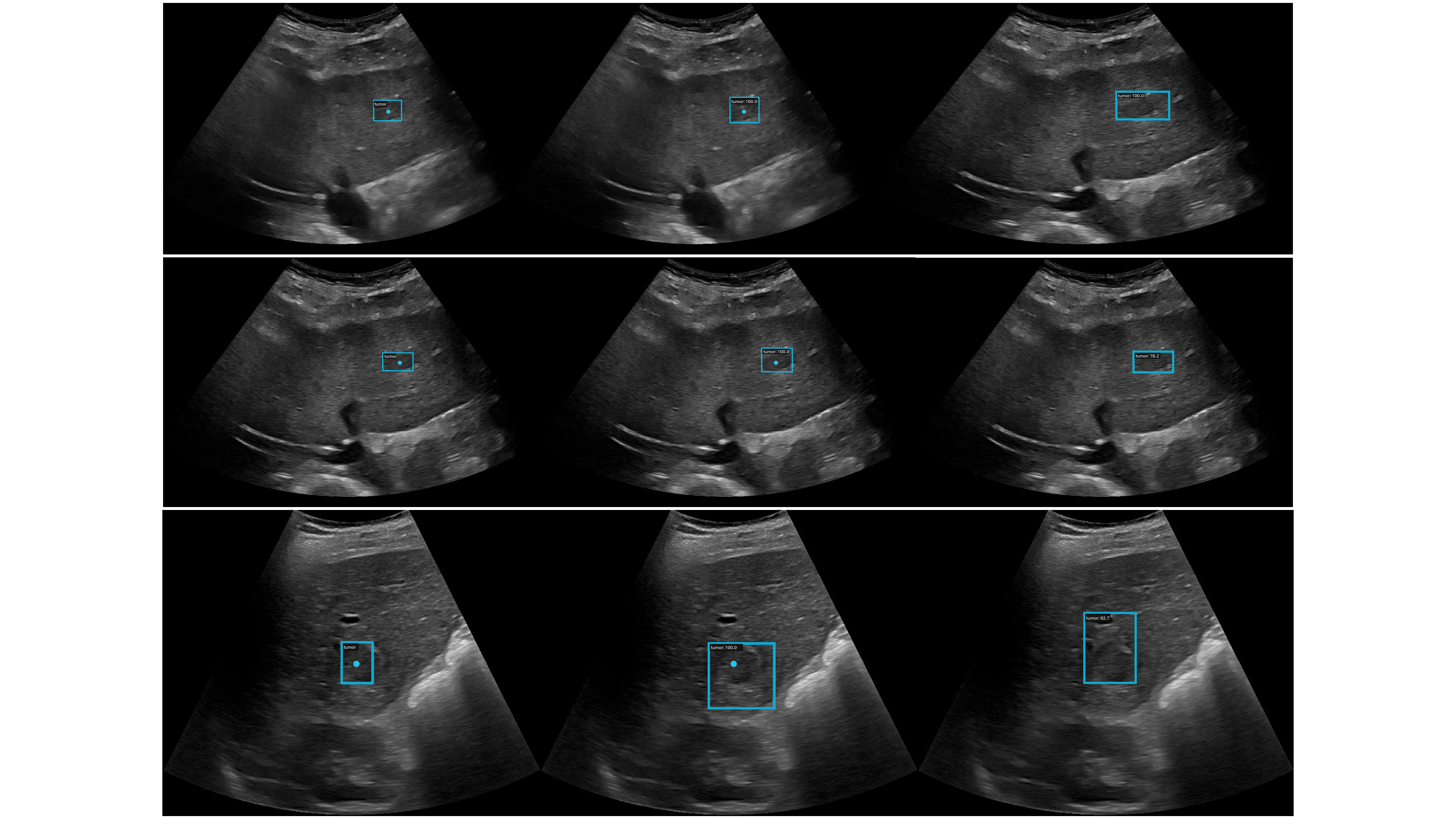}
\caption{Qualitative results of $S_{OTF}$ are presented. The left column displays the ground truth, featuring both point and corresponding bounding box annotations for lesions. The middle column depicts the predicted pseudo-labels with $S_{OTF}$, and the right column the prediction from the fully supervised model.}\label{quali}
\end{figure}

In Fig.\ref{quali}, we showcase qualitative results of our study. The left column displays the ground truth, featuring both point and corresponding bounding box annotations for lesions. The middle column depicts the predicted pseudo-labels with $S_{OTF}$. The right column displays predictions from the fully-supervised model. The first two rows display instances where accurate pseudo-labels were generated. However, the last row reveals a case where the model failed, incorrectly interpreting the ultrasound artifact, known as acoustic shadowing, as the periphery of the lesion. As mentioned, the OTF point is consistently localized on the near-center of the lesions.

Our method primarily focuses on annotation, making it inherently adaptable and compatible with various optimization strategies, such as self-supervised learning and active learning. Active learning streamlines the training of models by strategically selecting a subset of unlabeled data. This method focuses on choosing samples that, once annotated, contribute most effectively to the model's performance. This iterative process of model improvement aims to achieve high accuracy with fewer labeled instances, which is valuable when data annotation is expensive or time-consuming. \cite{kim2023active,wang2016cost} focus on identifying and annotating the most uncertain or challenging samples, thereby optimizing the learning process and efficiently utilizing the annotation budget. Those approaches, which emphasizes learning from complex cases, can be integrated with on-the-fly annotation strategies to allows for effective decision-making about which videos to annotate in-depth (box-annotation) and which to annotate weakly, ensuring a targeted and resource-efficient learning process.

\thispagestyle{default}
\section{Conclusion}\label{conclusion}

In this paper, we introduce an on-the-fly point annotation pipeline, enabling live video annotation to mitigate the tedious efforts specifically associated with video annotation. Every frame in the video is weakly-annotated, ensuring expert guidance throughout the process. Our method proves to allow precise tracking of structures, which can enable useful pseudo-labels generation compatible with weakly semi-supervised object detection pipelines, outperforming conventional annotation method at equivalent annotation budgets. A notable reduction in annotation costs is observed through the utilization of this strategy. The findings of this study underscore the need for optimization of video annotation processes, enabling the development of high-quality datasets. This approach fosters a more efficient utilization of expert resources, optimizing the balance between annotation accuracy and cost-effectiveness in medical imaging studies.
\section{Acknowledgement}
This research was conducted within the framework of the APEUS and TheraHCC 2.0 projects, which are supported by the ARC Foundation (www.fondation-arc.org). This work was also partially supported by French state funds managed within the 'Plan Investissements d’Avenir', funded by the ANR (reference ANR-10-IAHU-02 and ANR-21-RHUS-0001 DELIVER). This work was performed using HPC resources from GENCI–IDRIS (Grant 2023-AD011013698R1).

\noindent{\bf Ethical Approval:} This article does not contain any studies with human participants or animals performed by any of the authors.

\noindent{\bf Competing Interests:} The authors declare no conflict of interest.

\noindent{\bf Informed Consent:} This manuscript does not contain any patient data.

\bibliographystyle{sn-basic}
\bibliography{sn-bibliography}

\end{document}